\pdfoutput=1

\documentclass[11pt]{article}

\usepackage{emnlp2021}

\usepackage{times}
\usepackage{latexsym}

\usepackage[T1]{fontenc}

\usepackage[utf8]{inputenc}

\usepackage{microtype}

\usepackage{amsmath}
\usepackage{amsfonts}
\usepackage{subfigure}
\usepackage{graphicx}
\usepackage{booktabs}
\usepackage{url}
\usepackage{tabularx}
\usepackage{bigstrut}
\usepackage{multirow}
\usepackage{color}

\usepackage{enumitem}

\title{LayoutReader: Pre-training of Text and Layout for \\ Reading Order Detection}

\author{Zilong Wang$^{1}\thanks{Contributions during internship at MSRA.}$, Yiheng Xu$^{2*}$, Lei Cui$^{2}$, Jingbo Shang$^{1}$, Furu Wei$^{2}$ \\

$^{1}$University of California, San Diego\\
$^{2}$Microsoft Research Asia\\
\texttt{\{zlwang,jshang\}@ucsd.edu} \\
\texttt{\{t-yihengxu,lecu,fuwei\}@microsoft.com} \\
}

\date{}

\begin{document}
\maketitle
\begin{abstract}
Reading order detection is the cornerstone to understanding visually-rich documents (e.g., receipts and forms). Unfortunately, no existing work took advantage of advanced deep learning models because it is too laborious to annotate a large enough dataset. We observe that the reading order of WORD documents is embedded in their XML metadata; meanwhile, it is easy to convert WORD documents to PDFs or images. Therefore, in an automated manner, we construct \textbf{ReadingBank}, a benchmark dataset that contains reading order, text, and layout information for 500,000 document images covering a wide spectrum of document types. This first-ever large-scale dataset unleashes the power of deep neural networks for reading order detection. Specifically, our proposed \textbf{LayoutReader} captures the text and layout information for reading order prediction using the seq2seq model. It performs almost perfectly in reading order detection and significantly improves both open-source and commercial OCR engines in ordering text lines in their results in our experiments. We will release the dataset and model at \url{https://aka.ms/layoutreader}.


\end{abstract}

\section{Introduction}
\begin{figure*}[htbp]
\centering
\subfigure[]{
\includegraphics[width=0.20\linewidth]{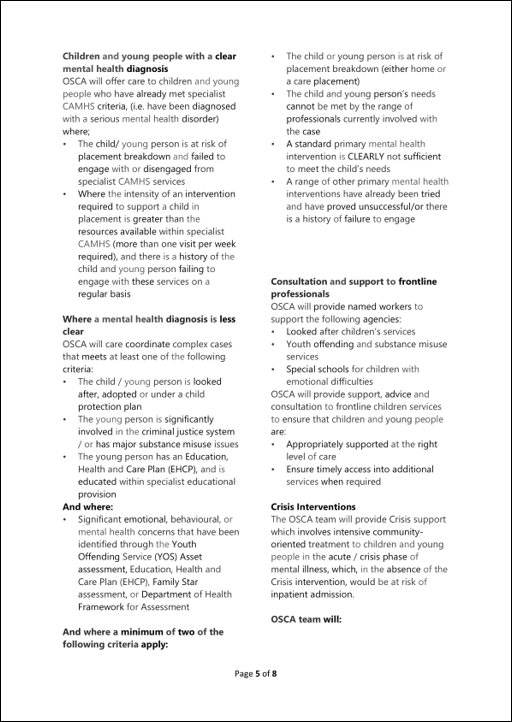}
}
\subfigure[]{
\includegraphics[width=0.20\linewidth]{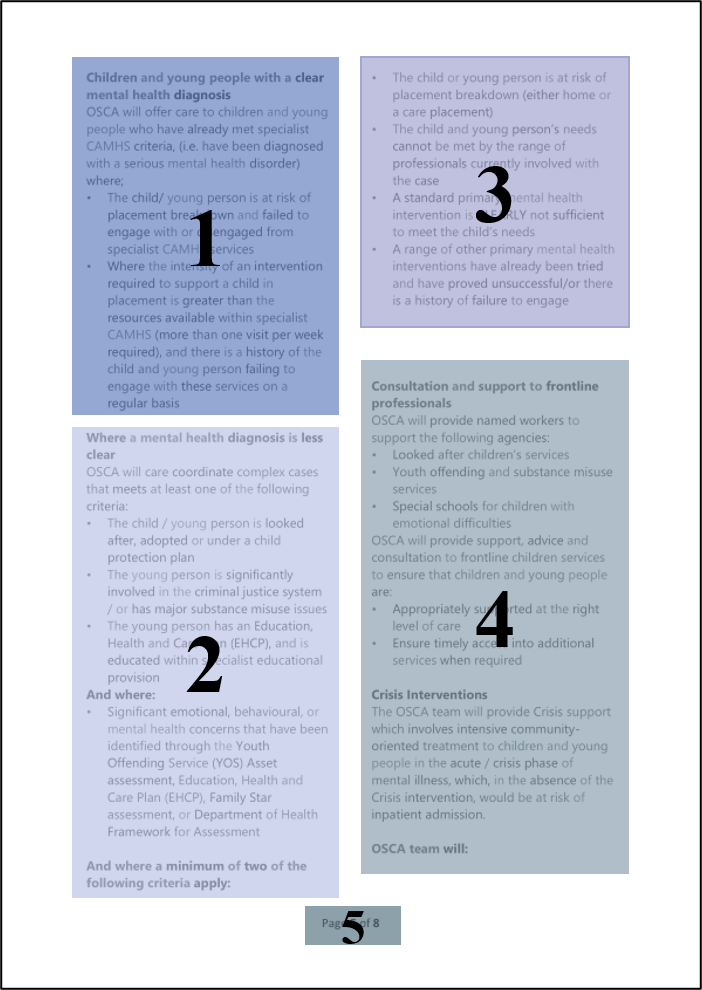}
}
\subfigure[]{
\includegraphics[width=0.20\linewidth]{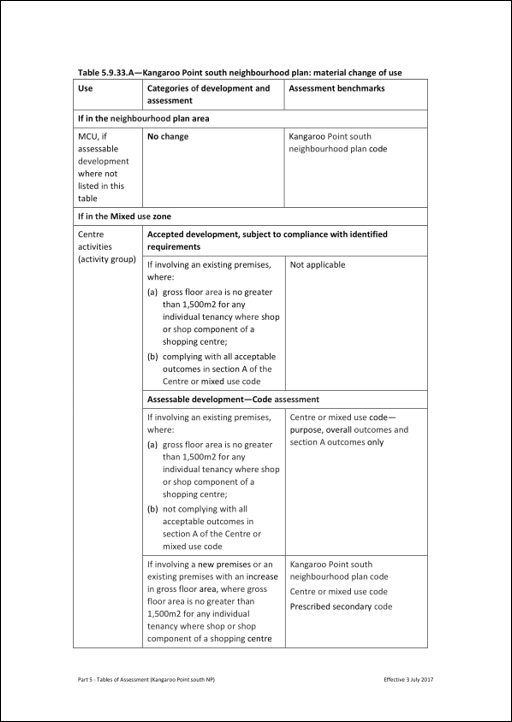}
}
\subfigure[]{
\includegraphics[width=0.20\linewidth]{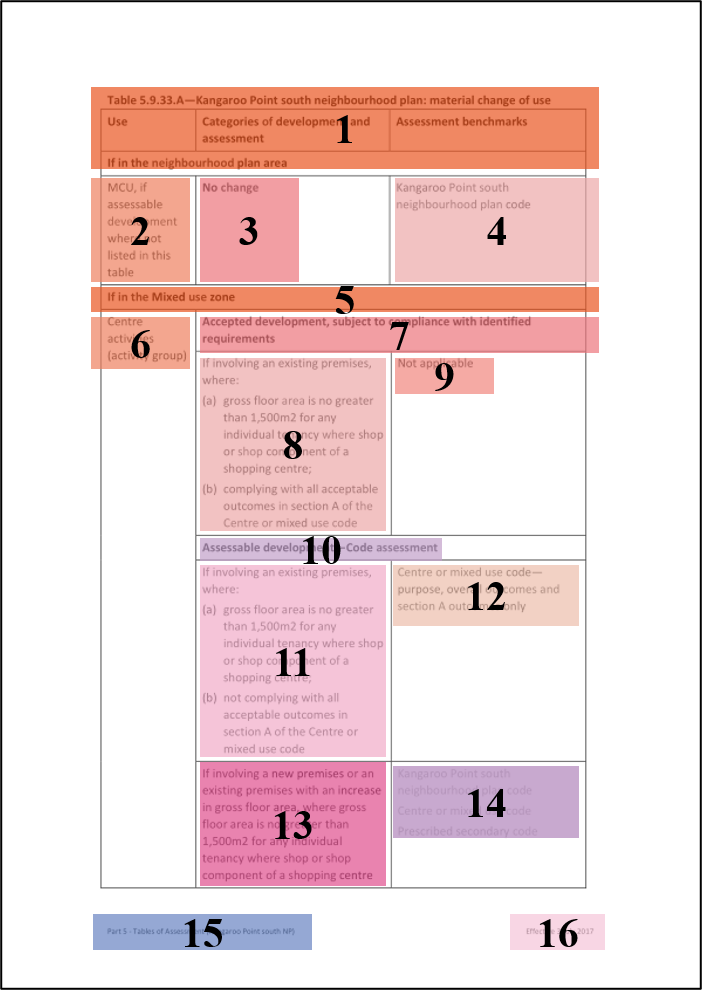}
}
\caption{Document image examples in ReadingBank with the reading order information. The colored areas show the paragraph-level reading order.}
\label{fig:1}
\end{figure*}

Reading order detection, aiming to capture the word sequence which can be naturally comprehended by human readers, is a fundamental task for visually-rich document understanding. 
Current off-the-shelf methods usually directly borrow the results from the Optical Character Recognition (OCR) engines \cite{10.1145/3394486.3403172} while most OCR engines arrange the recognized tokens or text lines in a top-to-bottom and left-to-right way~\cite{6628706}.
Apparently, as shown in Figure~\ref{fig:1}, this heuristic is not optimal for certain document types, such as multi-column templates, forms, invoices, and many others. 
An incorrect reading order will lead to unacceptable results for document understanding tasks such as the information extraction from receipts/invoices. Therefore, an accurate reading order detection model is indispensable to the document understanding tasks.



In the past decades,  some conventional machine learning based or rule based methods~\cite{aiello2003bidimensional,4377050,malerba2007learning,malerba2008machine,10.1145/2644866.2644883} have been proposed.
However, 
these approaches are usually trained with only a small number of samples within a restricted domain or resort to unsupervised methods with empirical rules, because it is too laborious to annotate a large enough dataset.
These models can barely show case studies of certain reading order scenarios and cannot be easily adapted for real-world reading order problems. 
Recently, deep learning models \cite{10.1007/978-3-030-58595-2_6} have been applied to address the reading order issues for images from E-commerce platforms. Although good performance has been achieved, it is time-consuming and labor-intensive to produce an in-house dataset, while they are still not publicly available to compare with other deep learning approaches. Therefore, to facilitate the long-term research of reading order detection, it is inevitable to leverage automated approaches to create a real-world dataset in general domains, not only with high quality but also of larger magnitude than the existing datasets.


To this end, we propose ReadingBank, a benchmark dataset with 500,000 real-world document images for reading order detection. 
Distinct from the conventional human-labeled data, the proposed method obtains high-quality reading order annotations in a simple but effective way with automated metadata extraction. 
Inspired by existing document layout annotations~\cite{Siegel2018ExtractingSF, Zhong2019PubLayNetLD, li2020tablebank, li2020docbank}, there are a large number of Microsoft WORD documents with a wide variety of templates that are available on the internet. 
Typically, the WORD documents have two formats: the binary format (Doc files) and the XML format (DocX files). 
In this work, we exclusively use WORD documents with the XML format
as the reading order information is embedded in the XML metadata.
Furthermore, we convert the WORD documents into the PDF format so that the 2D bounding box of each word can be easily extracted using any off-the-shelf PDF parser. 
Finally, we apply a carefully designed coloring scheme to align the text in the XML metadata with the bounding boxes in PDFs. 

With the large-scale dataset, it is possible to take advantage of deep neural networks to solve reading order detection task. 
We further propose LayoutReader, a novel reading order detection model in which the seq2seq model is used by encoding the text and layout information and generating the index sequence in the reading order. Ablation studies on the input modalities show that both text and layout information are essential to the final performance. The LayoutReader with both modalities surpasses other comparative methods and performs almost perfectly in reading order detection.
In addition, we also adapt the results of LayoutReader to open-source and commercial OCR engines in ordering text lines. 
Experiments show that the line ordering of both open-source and commercial OCR engines can be greatly improved. 
We believe that ReadingBank and LayoutReader will empower more deep learning models in the reading order detection task and foster more customized neural architectures to push the new SOTA on this task.

The contributions are summarized as follows:

\begin{itemize}[nosep, leftmargin=*]
    \item We present ReadingBank, a benchmark dataset with 500,000 document images for reading order detection. To the best of our knowledge, this is the first large-scale benchmark for the research of reading order detection.
    \item We propose LayoutReader for reading order detection and conduct experiments with different parameter settings. The results confirm the effectiveness of LayoutReader in detecting reading order of documents and improving line ordering of OCR engines.
    \item The ReadingBank dataset and LayoutReader models will be publicly available to support more deep learning models on reading order detection.
\end{itemize}

\noindent\textbf{Reproducibility.} We will release the code and datasets at \url{https://aka.ms/layoutreader}.

\section{Problem Formulation}
Reading order refers to a well-organized readable word sequence. Although it seems a fundamental requirement of NLP datasets, it is non-trivial to obtain proper reading orders from document images due to various formats, e.g., tables, multiple columns, and most OCR engines fail to provide the proper reading order.

To solve this problem, we address the reading order detection task, aiming to extract the natural reading sequence from document images.  Specifically, given a visually-rich document image $\mathcal{D}$, we acquire discrete token set $\{t_1, t_2, t_3,...\}$ where each token $t_i$ consists of a word $w_i$ and the its bounding box coordinates $(x_0^i,y_0^i, x_1^i, y_1^i)$ (the left-top corner and right-bottom corner). 
Equipped with the textual and layout information of the tokens in the document image, we intend to sort the tokens into the reading order.

\begin{figure}[t]
    \centering
    \includegraphics[width=0.9\linewidth]{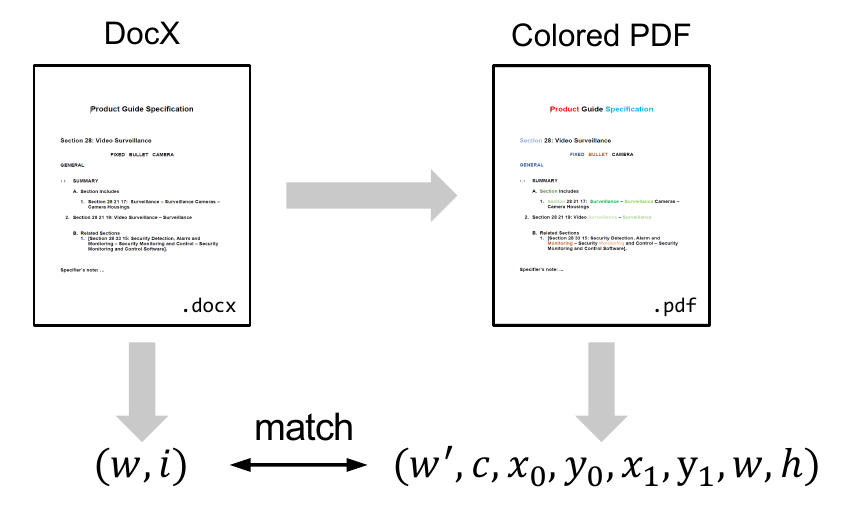}
    \caption{Building pipeline of ReadingBank, where $(w,i)$ is the pair of word and its appearance index and $(w^{\prime}, c, x_0, y_0, x_1, y_1, w, h)$ is the word, word color and layout information.}
    \label{fig:doc_proc}
\end{figure}

\section{ReadingBank}
ReadingBank includes two parts, the word sequence and its corresponding bounding box coordinates. We denote the word sequence as Reading Sequence that is extracted from DocX files. The corresponding bounding boxes are extracted from the PDF files which are generated from DocX files. We propose a coloring scheme to solve the word duplication when we match each word and its bounding box.

In this section, we introduce the data pipeline in detail, including document collection, reading sequence extraction, and layout alignment with the coloring scheme. The current ReadingBank totally includes 500,000 document pages, where the training set includes 400,000 document pages and both the validation set and the test set include 50,000 document pages, respectively.

\subsection{Document Collection}
We crawl the WORD documents in DocX format from the internet considering the robots exclusion standard as well as the public domain license. \footnote{More ethical details are included in the Ethical Consideration section.} 
We further use the language detection API \footnote{\url{https://azure.microsoft.com/en-us/services/cognitive-services/text-analytics/}} with a high confidence threshold to filter non-English or bilingual documents because we focus on the reading order detection for English documents in this work. The reading order detection of other languages will be our future work. We only keep the pages with more than 50 words to guarantee the enough information on each page. In this way, we have totally collected 210,000 WORD documents in English and each page in the documents is informative enough. We further randomly select 500,000 pages to build our dataset.

\begin{table*}[t]
\small
  \centering
\begin{tabular}{ccccccc}
\toprule
\multirow{2}[2]{*}{\textbf{Split}} & \multirow{2}[2]{*}{\textbf{\#Word Avg.}} & \multirow{2}[2]{*}{\textbf{Avg. BLEU}} & \multicolumn{4}{c}{\textbf{BLEU Distribution}} \bigstrut[t]\\
      &       &       & \textbf{(0.00, 0.25]} & \textbf{(0.25, 0.50]} & \textbf{(0.50, 0.75]} & \textbf{(0.75, 1.00]} \bigstrut[b]\\
\midrule
\multirow{2}[2]{*}{Train} & \multirow{2}[2]{*}{196.38} & \multirow{2}[2]{*}{0.6974} & 9,666  & 58,785 & 155,662 & 175,884 \bigstrut[t]\\
      &       &       & 2.42\% & 14.70\% & 38.92\% & 43.97\% \bigstrut[b]\\
\hline
\multirow{2}[2]{*}{Validation} & \multirow{2}[2]{*}{196.02} & \multirow{2}[2]{*}{0.6974} & 1,203  & 7,351  & 19,387 & 22,053 \bigstrut[t]\\
      &       &       & 2.41\% & 14.70\% & 38.78\% & 44.11\% \bigstrut[b]\\
\hline
\multirow{2}[2]{*}{Test} & \multirow{2}[2]{*}{196.55} & \multirow{2}[2]{*}{0.6972} & 1,232  & 7,329  & 19,555 & 21,893 \bigstrut[t]\\
      &       &       & 2.46\% & 14.66\% & 39.10\% & 43.78\% \bigstrut[b]\\
\hline
\multirow{2}[2]{*}{All} & \multirow{2}[2]{*}{196.36} & \multirow{2}[2]{*}{0.6974} & 12,101 & 73,465 & 194,604 & 219,830 \bigstrut[t]\\
      &       &       & 2.42\% & 14.69\% & 38.92\% & 43.97\% \bigstrut[b]\\
\bottomrule
\end{tabular}%
  \caption{Dataset statistics of training, validation, and test sets in ReadingBank. The BLEU scores are calculated for the left-to-right and top-to-bottom order to measure the difficulty of training samples}
  \label{tab:statistics}%
\end{table*}%

\subsection{Reading Sequence Extraction}
The reading order in ReadingBank refers to the order of words in the DocX files. Each DocX file is a compressed archive where its word sequence can be parsed from its internal Office XML code. We adopt an open source tool \verb|python-docx|\footnote{\url{https://pypi.org/project/python-docx/}} to parse the DocX file and extract the word sequence from the XML metadata. The tool also enables us to change the words' color for the layout alignment step.

We first extract the paragraphs and the tables sequentially from the parsing result. Then we traverse the paragraphs line by line and the tables cell by cell and obtain the word sequence in the DocX file. We denote the sequence as $[w_1, w_2, ..., w_n]$, where $n$ is the number of words in this document. The obtained sequence is the reading order without the layout information and is denoted as the Reading Sequence. We would align the bounding box to each word in this sequence in the following steps.

\subsection{Layout Alignment with Coloring Scheme}
In our extensive collection, the same word may appear multiple times in the same document, and we need to solve this duplication when we assign the coordinates to each word.
Therefore, we give each word an extra label indicating its appearance index. For example, given a sequence [the, car, hits, the, bus], the extra labels should be [0, 0, 0, 1, 0] since there are two ``the''s in this example. In this way, each pair of the word and its appearance index is unique and can serve as the key when assigning the location coordinates. 

Meanwhile, we propose the coloring scheme to show the keys in the DocX file without changing the original layout pattern. We map the appearance index to the RGB colors through $\mathcal{C}: \mathbb{N} \mapsto \mathbf{RGB}$ and color the words accordingly. To eliminate the interference from the original word color, we first color all the words into black.
\begin{align*}
    r &= i \& 0\text{x}110000 \\
    g &= i \& 0\text{x}001100 \\
    b &= i \& 0\text{x}000011 \\
    \mathcal{C}(i) &= (\mathbf{R}: r,\mathbf{G}: g, \mathbf{B}: b)
\end{align*}
where $i$ is the appearance index of the given word; $\&$ is the bit-wise and operation; $\mathcal{C}$ is the mapping function.

Although DocX files provide a reasonable reading sequence but the location of each word in DocX files is not fixed. Therefore, we use the PDF files produced by the colored DocX files as an intermediate to extract layout information. We adopt \verb|PDF Metamorphosis .Net|\footnote{\url{https://sautinsoft.com/products/pdf-metamorphosis/}} to convert the DocX files to PDF and use an open source tool \verb|MuPDF|\footnote{\url{https://www.mupdf.com/}} as the PDF parser. We extract the words, bounding box coordinates, word color from the PDF file. Since the mapping function $\mathcal{C}$ is a one-to-one correspondence, we easily get the appearance index by using the coloring scheme. For the convenience of future study, we also extract the height and width of the page. 
In this way, we can build a one-to-one matching between the Reading Sequence and the PDF layout information.
\begin{align*}
    (w, i) 	&\leftrightarrow (w^{\prime}, c, x_0, y_0, x_1, y_1, W, H) \notag \\
\text{subject to } &w = w^{\prime} \text{; } c = \mathcal{C}(i) 
\end{align*}
where $w$ and $w^{\prime}$ are the word in DocX and PDF, respectively; $i$ is the appearance index of $w$; $c$ is the word color recognized by PDF parser; $x_0$, $y_0$, $x_1$, $y_1$ are the left-top and right-bottom coordinates; $W$, $H$ are the width and height of the page where the word locates. In the post-processing stage, we collect data for each page and build our dataset. 

\subsection{Dataset Statistics}
The ReadingBank consists of 500,000 document pages including the image and the sequence of words and coordinates in reading order. We divide the whole dataset by ratio 8:1:1 for training, validation, and testing. Table \ref{tab:statistics} shows the details of the three subsets. The average word number, the average sentence-level BLEU score and the sentence-level BLEU score distribution are reported. The BLEU scores are calculated for the left-to-right and top-to-bottom order using the groundtruth reading order as the reference, so as to measure the difficulty of training samples.
To guarantee the data balance, the distribution of word number and BLEU score are consistent as we randomly gather pages into each subset. We assume the ReadingBank will not suffer from the data unbalance during pre-training or fine-tuning.

\begin{figure*}[t]
    \centering
    \includegraphics[width=0.9\linewidth]{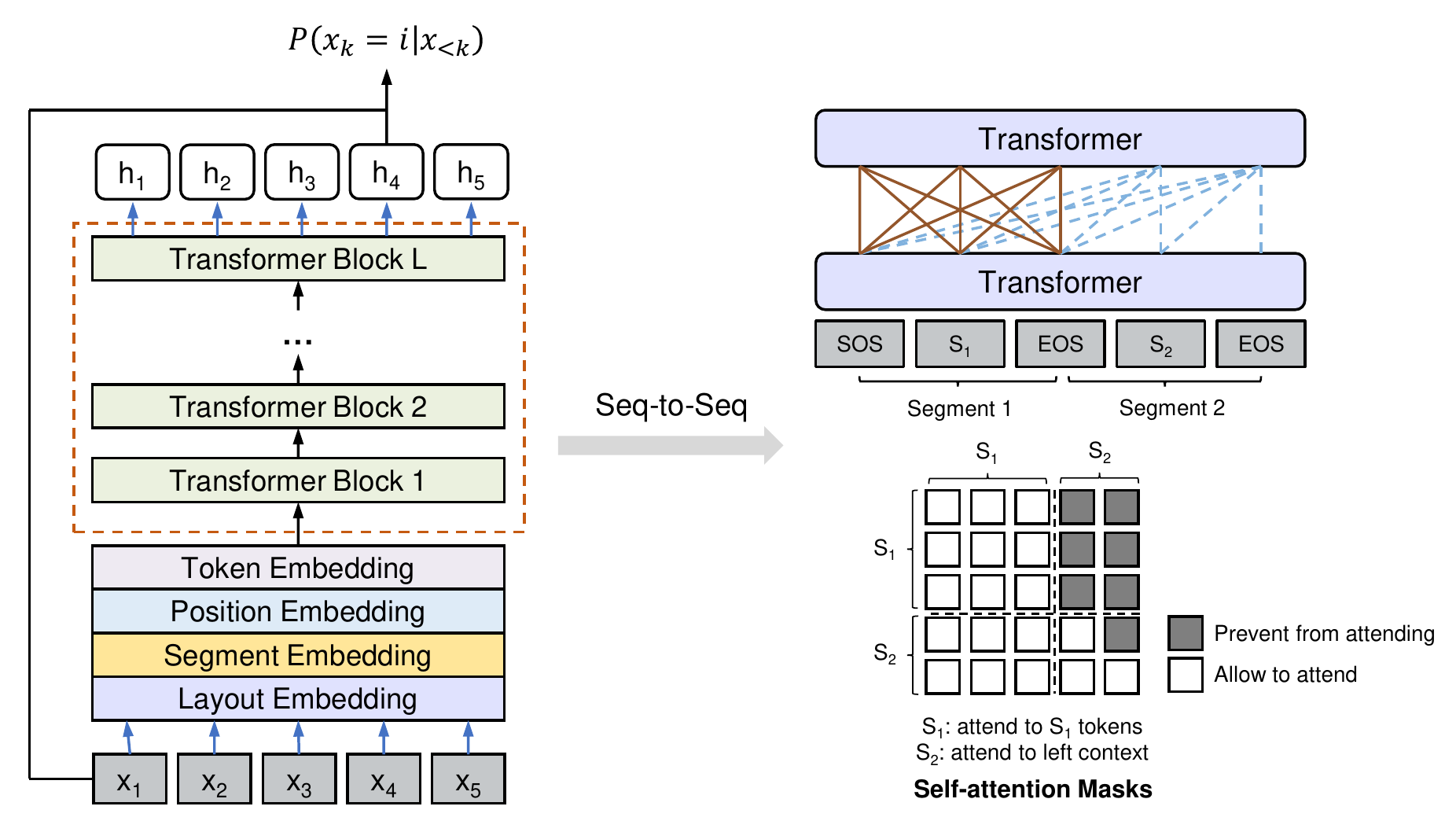}
    \caption{LayoutReader architecture for the reading order detection. The self-attention is designed for sequence-to-sequence modeling and the generation step is modified to predict the indices in the source segment.}
    \label{fig:seq2seq}
\end{figure*}

\section{LayoutReader}
With ReadingBank, we further propose LayoutReader to solve the reading order detection task. LayoutReader is a sequence-to-sequence model using both textual and layout information, where we leverage the layout-aware language model LayoutLM~\cite{10.1145/3394486.3403172} as encoder and modify the generation step in the encoder-decoder structure to generate the reading order sequence. 

\paragraph{Encoder:} In the encoding stage, LayoutReader packs the pair of source and target segments into a contiguous input sequence of LayoutLM and carefully designs the self-attention mask to control the visibility between tokens. As shown in Figure \ref{fig:seq2seq}, LayoutReader allows the tokens in the source segment to attend to each other while preventing the tokens in the target segment from attending to the rightward context. If 1 means allowing and 0 means preventing, the detail of the mask $M$ is as follows:
\begin{align*}
    M_{i,j} = 
    \begin{cases} 
        1,  & \mbox{if }i < j \mbox{ or } i,j \in \mbox{src}\\
        0, & \mbox{otherwise}
    \end{cases}
\end{align*}
where $i$, $j$ are the indices in the packed input sequence, so they may be from source or target segments; $i,j \in \mbox{src}$ means both tokens are from source segment.



\paragraph{Decoder:} In the decoding stage, since the source and target are reordered sequences, the prediction candidates can be constrained to the source segment. Therefore, we ask the model to predict the indices in the source sequence. The probability is calculated as follows:
\begin{align*}
    \mathcal{P}(x_k=i|x_{<k}) = \frac{\exp{(e_i^Th_k + b_k)}}{\sum_j \exp{(e_j^Th_k + b_k)}}
\end{align*}
where $i$ is an index in the source segment; $e_i$ and $e_j$ are the i-th and j-th input embeddings of the source segment; $h_k$ is the hidden states at the k-th time step; $b_k$ is the bias at the k-th time step.

\begin{table*}[t]
\small
  \centering
\begin{tabular}{cccc}
\hline
\textbf{Method} & \textbf{Encoder} & \textbf{Avg. Page-level BLEU ↑} & \textbf{ARD ↓} \bigstrut\\
\hline
 Heuristic Method & -     & 0.6972 & 8.46 \bigstrut\\
\hline
\multirow{2}[2]{*}{LayoutReader (text only)} & BERT  & 0.8510 & 12.08 \bigstrut[t]\\
      & UniLM & 0.8765 & 10.65 \bigstrut[b]\\
\hline
LayoutReader (layout only) & LayoutLM (layout only) & 0.9732 & 2.31 \bigstrut\\
\hline
LayoutReader & LayoutLM & \textbf{0.9819} & \textbf{1.75} \bigstrut\\
\hline
\end{tabular}%
    \caption{Evaluation results of the LayoutReader on the reading order detection task, where the source-side of training/testing data is in the left-to-right and top-to-bottom order}
      \label{tab:results}
\end{table*}%

\section{Experiments}
We introduce the comparative methods, implementation details, and evaluation metrics for the experiments. We design three experiments for LayoutReader on ReadingBank, including reading order detection, input order study, and adaption on OCR engines. In addition, we also show the real-world examples in the case study. 

\subsection{Comparative Methods}
LayoutReader considers both text and layout information with the multi-modal encoder LayoutLM. To further study the role of each modality, we design two comparative models, including LayoutReader (layout only) and LayoutReader (text only). We also report the results of the Heuristic Method as our baseline.
\paragraph{Heuristic Method:} This method refers to sorting words from left to right and from top to bottom. 

\paragraph{LayoutReader (text only):} We replace LayoutLM with textual language models, e.g. BERT~\cite{devlin2018bert}, UniLM~\cite{dong2019unified}, which means LayoutReader (text only) predicts the reading order only through textual information. Our experiments build two versions of LayoutReader (text only), which use BERT or UniLM as a substitute of LayoutLM, respectively.
\paragraph{LayoutReader (layout only):} We remove the token embeddings in LayoutLM. The token embeddings are vital for Transformer to extract textual information. After removing these embeddings, LayoutReader (layout only) only considers the 1D and 2D positional layout information.

\subsection{Implementation Details}
Our implementation is built upon the HuggingFace Transformers~\cite{wolf2019huggingface} and the LayoutReader is implemented with the s2s-ft toolkit from the repository of \citet{dong2019unified}\footnote{\url{https://github.com/microsoft/unilm/tree/master/s2s-ft}}. The pre-trained models used are in their base version. We use 4 Tesla V100 GPUs with batch size of 4 per GPU during training. The number of training epochs is 3 and the training process takes approximately 6 hours.
We optimize the models with the AdamW optimizer. The initial learning rate is $7\times 10^{-5}$ and the number of warm-up steps is $500$.

\subsection{Evaluation Metrics}

\paragraph{Average Page-level BLEU:} The BLEU score~\cite{papineni2002bleu} is widely used in sequence generation. Since LayoutReader is built on a sequence-to-sequence model, it is natural to evaluate our models with BLEU scores. BLEU scores measure the n-gram overlaps between the hypothesis and reference. We report Average Page-level BLEU in our experiments. The page-level BLEU refers to the micro-average precision of n-gram overlaps within a page.

\paragraph{Average Relative Distance (ARD):} The ARD score is proposed to evaluate the difference between reordered sequences. It measures the relative distance between the common elements in the different sequence. Since our reordered sequence is generated, the ARD allows the element omission but adds a punishment for it. Given a sequence $A=[e_1,e_2,...,e_n]$ and its generated reordered sequence $B=[e_{i_1},e_{i_2},...,e_{i_m}]$, where $\{i_1,i_2,...,i_m\}\subseteq \{1,2,...,n\}$, the ARD score is calculated as follows:
\begin{align*}
    &s(e_k, B) = 
\begin{cases} 
|k - I(e_k, B)|,  & \mbox{if }e_k \in B\\
n, & \mbox{otherwise}
\end{cases}  \label{ard}\\
&\text{ARD}(A, B) = \frac{1}{n} \sum_{e_k\in A} s(e_k, B)
\end{align*}
where $e_k$ is the k-th element in sequence $A$; $I(e_k, B)$ is the index of $e_k$ in sequence $B$; $n$ is the length of sequence A.

\begin{table*}[t]
\small
  \centering
\begin{tabular}{ccccccc}
\toprule
\multirow{2}[2]{*}{\textbf{Method}} & \multicolumn{3}{c}{\textbf{Avg. Page-level BLEU ↑}} & \multicolumn{3}{c}{\textbf{ARD ↓}} \bigstrut[t]\\
      & \textbf{$r$=100\%} & \textbf{$r$=50\%} & \textbf{$r$=0\%} & \textbf{$r$=100\%} & \textbf{$r$=50\%} & \textbf{$r$=0\%} \bigstrut[b]\\
\midrule
LayoutReader (text only, BERT) & 0.3355 & 0.8397 & 0.8510 & 77.97 & 15.62 & 12.08 \bigstrut[t]\\
LayoutReader (text only, UniLM) & 0.3440 & 0.8588 & 0.8765 & 78.67 & 13.65 & 10.65 \bigstrut[b]\\
\hline
LayoutReader (layout only) & 0.9701 & 0.9729 & 0.9732 & 2.85  & 2.61  & 2.31 \bigstrut\\
\hline
LayoutReader & \textbf{0.9765} & \textbf{0.9788} & \textbf{0.9819} & \textbf{2.50} & \textbf{2.24} & \textbf{1.75} \bigstrut\\
\bottomrule
\end{tabular}%

\caption{Input order study with left-to-right and top-to-bottom inputs in evaluation, where $r$ is the proportion of shuffled samples in training.}
  \label{tab:confusion_heuristic_input}%
\end{table*}%

\begin{table*}[t]
\small
  \centering
\begin{tabular}{ccccccc}
\toprule
\multirow{2}[2]{*}{\textbf{Method}} & \multicolumn{3}{c}{\textbf{Avg. Page-level BLEU ↑}} & \multicolumn{3}{c}{\textbf{ARD ↓}} \bigstrut[t]\\
      & \textbf{$r$=100\%} & \textbf{$r$=50\%} & \textbf{$r$=0\%} & \textbf{$r$=100\%} & \textbf{$r$=50\%} & \textbf{$r$=0\%} \bigstrut[b]\\
\midrule
LayoutReader (text only, BERT) & 0.3085 & 0.2730 & 0.1711 & 78.69 & 85.44 & \textbf{67.96} \bigstrut[t]\\
LayoutReader (text only, UniLM) & 0.3119 & 0.2855 & 0.1728 & 80.00 & 85.60 & 71.13 \\
LayoutReader (layout only) & 0.9718 & 0.9714 & 0.1331 & 2.72  & 2.82  & 105.40 \bigstrut[b]\\
\hline
LayoutReader & \textbf{0.9772} & \textbf{0.9770} & \textbf{0.1783} & \textbf{2.48} & \textbf{2.46} & 72.94 \bigstrut\\
\bottomrule
\end{tabular}%

    \caption{Input order study with token-shuffled inputs in evaluation, where $r$ is the proportion of shuffled samples in training.}
      \label{tab:confusion_random_input}%
\end{table*}%

\subsection{Reading Order Detection}
We train the models with left-to-right and top-to-bottom ordered inputs and report the evaluation results on the test set of ReadingBank in Table \ref{tab:results}. We also report the results of the heuristic method. The results show that LayoutReader is superior and achieves the SOTA results compared with other baselines. It improves the average page-level BLEU by 0.2847 and decreases the ARD by 6.71. Even if we remove some of the input modalities, there is still 0.16 and 0.27 improvements of BLEU in LayoutReader (text only) and LayoutReader (layout only), and there is a steady 6.15 reduction of ARD in LayoutReader (layout only). However, we also see a drop of ARD in LayoutReader (text only), mainly because of the severe punishment in ARD for token omission (see ARD definition). LayoutReader (text only) can guarantee the right order of tokens but suffers from the incompleteness of generation.
We also conclude that the layout information plays a more important role than textual information in the reading order detection. LayoutReader (layout only) surpasses the LayoutReader (text only) greatly by about 0.1 in BLEU and about 9.0 in ARD.


\subsection{Input Order for Training and Testing}
We shuffle the input tokens of sequence-to-sequence model in a certain proportion of training samples to study the accuracy of LayoutReader for different input orders. The proportion of token-shuffled training samples is denoted as $r$. We build three versions of comparative models with $r$ equaling $100\%$, $50\%$ and $0\%$. 
The left-to-right and top-to-bottom order provide remarkable hints for reading order detection. However, in this input order study, these hints are incomplete during training. We design two evaluation methods. Table \ref{tab:confusion_heuristic_input} shows the results when we evaluate the comparative models with left-to-right and top-to-bottom inputs. Table \ref{tab:confusion_random_input} shows the results when we evaluate the comparative models with token-shuffled inputs.

From Table \ref{tab:confusion_heuristic_input}, we observe that LayoutReader (layout only) and LayoutReader are more robust to the shuffled tokens during training, and all three comparative models perform well with the left-to-right and top-to-bottom inputs in evaluation. We attribute it to the consideration of layout information, which is consistent under shuffling.

From Table \ref{tab:confusion_random_input}, we see a drop when we train LayoutReader with $r=0\%$ token-shuffled inputs and evaluate it with all token-shuffled inputs. We explain that models trained on $r=0\%$ token-shuffled inputs tend to overfit the left-to-right and top-to-bottom order due to overlaps between this order and groundtruth, while the token-shuffled inputs in evaluation are totally unseen to these models.



\subsection{Adaption to OCR Engines} \label{sec:adation}
Most OCR engines provide reading order information for the text lines, where some of them may be problematic. To improve the text line ordering, we extend the token-level reading order to text lines and adapt it to OCR engines.

We first assign each token in our token-level order to the text lines according to the percentage of spatial overlapping. Given a token bounding box $b$ and a text line bounding box $B$, the token is assigned to the text line which overlaps the most with the token, i.e. $\hat{B} = \text{argmax}_{B} (B \cap b)$, where $\cap$ means spacial overlapping. Then we calculate the minimum of token indices in each text line as its ranking value and produce an improved text line order from the token-level order.
\begin{table}[t]
\small
  \centering
    \begin{tabular}{ccc}
    \toprule
    \textbf{Method} & 
           \textbf{Avg. Page-level BLEU ↑} & \textbf{ARD ↓} \bigstrut[b]\\
    \midrule
    Heuristic Method & 0.3391 & 13.61 \bigstrut\\
    \hline
    Tesseract OCR   & 0.7532 & 1.42 \bigstrut\\
    \hline
    LayoutReader & \textbf{0.9360} & \textbf{0.27} \bigstrut\\
    \bottomrule
    \end{tabular}%
      \caption{Adaption to text lines of Tesseract OCR}
  \label{tab:adaption_on_tesseract}%
\end{table}%

\begin{table}[t]
\small
  \centering
    \begin{tabular}{ccc}
    \toprule
    \textbf{Method} 
          & \textbf{Avg. Page-level BLEU ↑} & \textbf{ARD ↓} \bigstrut[b]\\
    \midrule
    Heuristic Method & 0.3752 & 10.17 \bigstrut\\
    \hline
    The commercial OCR & 0.8530 & 2.40 \bigstrut\\
    \hline
    LayoutReader & \textbf{0.9430} & \textbf{0.59} \bigstrut\\
    \bottomrule
    \end{tabular}%
      \caption{Adaption to text lines of the commercial OCR}
  \label{tab:adaption_on_ms}%
\end{table}%

\begin{figure*}[t]
\centering
\subfigure[Original image]{
\includegraphics[width=0.20\linewidth]{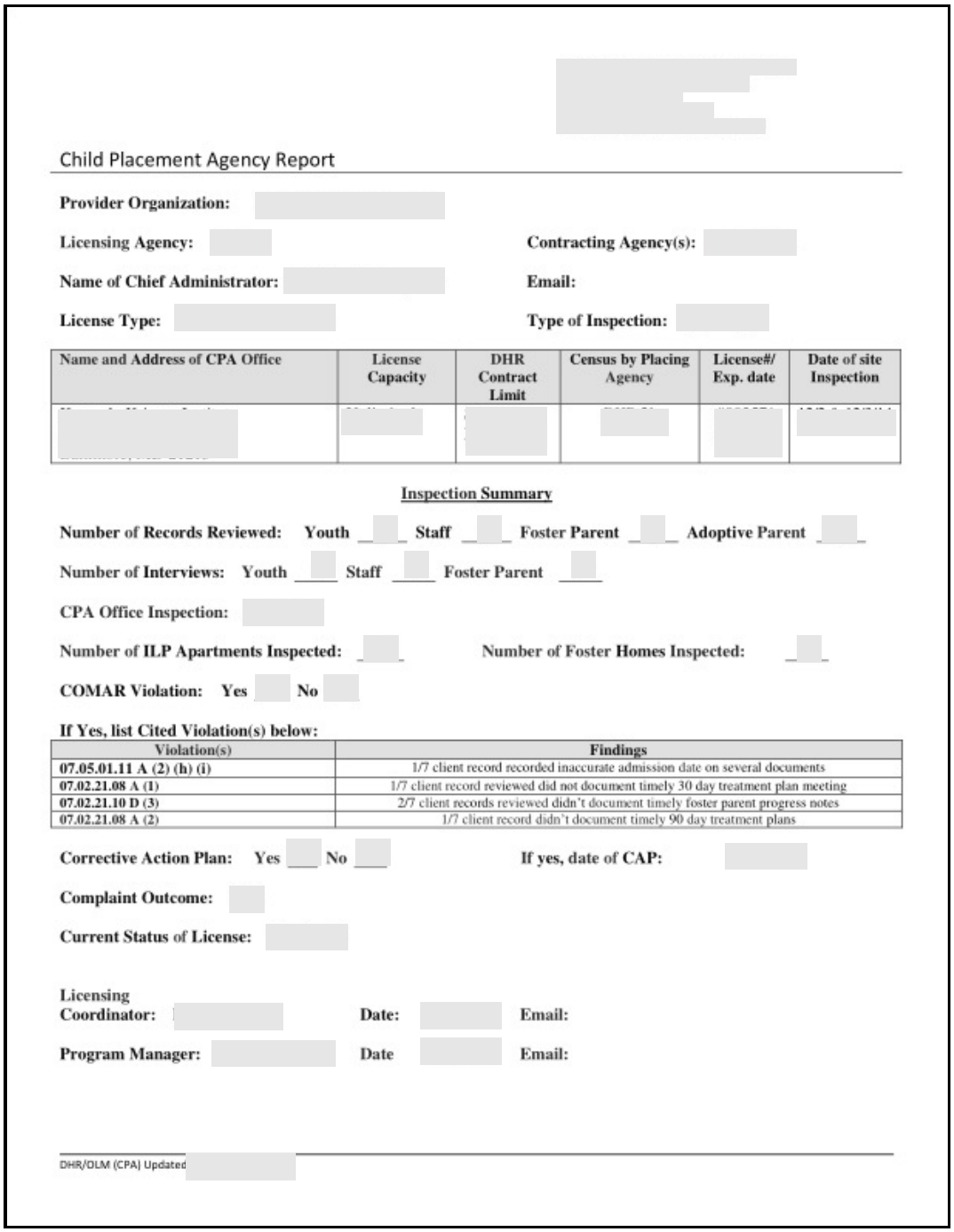}
}
\subfigure[Groundtruth]{
\includegraphics[width=0.20\linewidth]{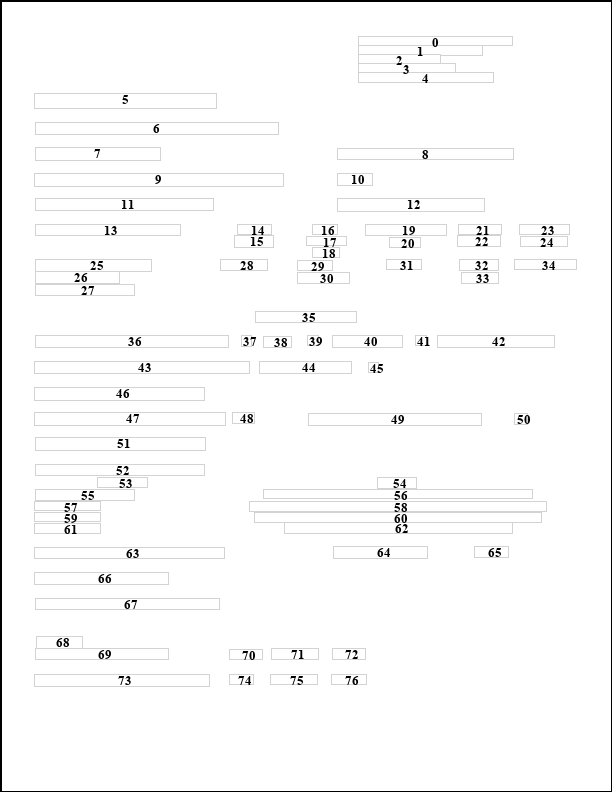}
}
\subfigure[The commercial OCR]{
\includegraphics[width=0.20\linewidth]{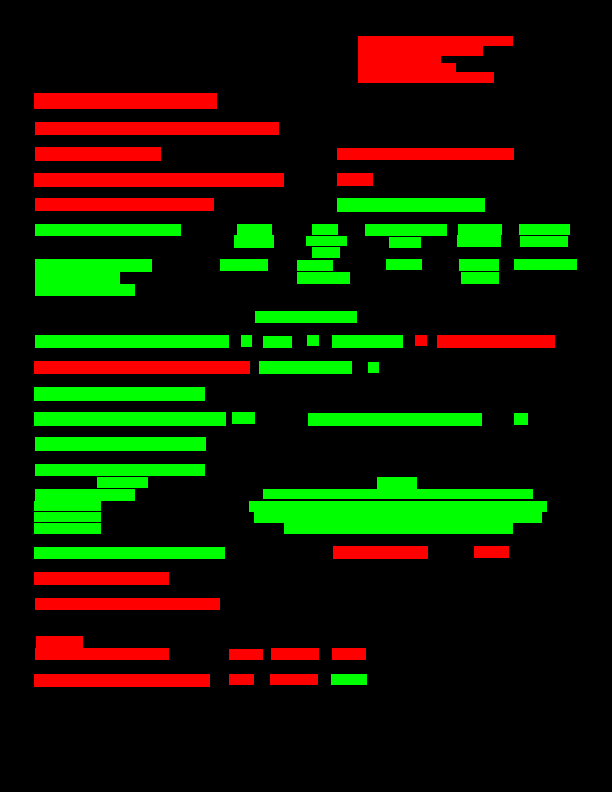}
\label{fig:case_study_ocr}
}
\subfigure[LayoutReader]{
\includegraphics[width=0.20\linewidth]{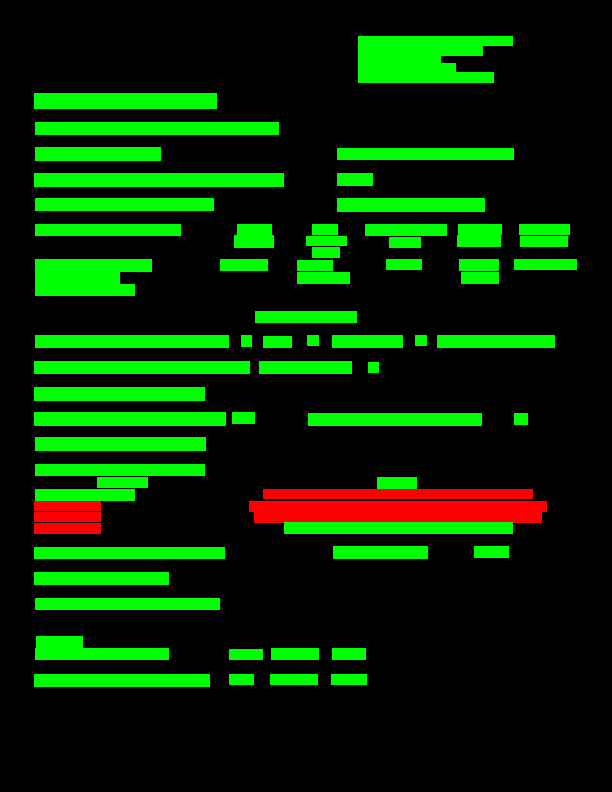}
\label{fig:case_study_layoutlm}
}
\caption{Case Study: (a) is the original image (some fields are masked because of privacy); (b) is the text line reading order groundtruth from ReadingBank Adaption; (c) and (d) are the results of a commercial OCR engine and LayoutReader Adaption where green and red denote the correct and incorrect predicted indices.}
\label{fig:case_study}
\end{figure*}

It should be noted that the token-level order can be the order given by ReadingBank or the result generated by LayoutReader. Therefore, we build a text line ordering groundtruth by adapting the ReadingBank to text lines and evaluate the performance of LayoutReader in text line ordering accordingly. We also report the performance of the Heuristic Method and OCR engines. We conduct experiments with two OCR engines, including an open source OCR engine Tesseract, and a cloud-based commercial OCR API. The results are shown in Table \ref{tab:adaption_on_tesseract} and Table \ref{tab:adaption_on_ms}. We can see a great improvement with LayoutReader Adaption. This experiment further demonstrates the effectiveness and extends the application of LayoutReader.


\subsection{Case Study}
We select a representative example from our test set and show the text line orders in Figure \ref{fig:case_study}. We compare the text line order of the commercial OCR engine and LayoutReader Adaption with the groundtruth from ReadingBank Adaption. We visualize the results with colors where green and red denotes correct and incorrect results. We see LayoutReader Adaption improves the text line ordering of the OCR engine, which is consistent with our results in Section \ref{sec:adation}.

\section{Related Work}

Reading order detection was first proposed in~\citep{aiello2003bidimensional}, where they used a propositional language of qualitative rectangle relations to detect reading order from document images. This is also considered as the first rule-based reading order detection system. With the development of machine learning methods,~\citep{4377050} proposed a probabilistic classifier using the Bayesian framework and reconstructing either single or multiple chains of layout components. Meanwhile, \citep{malerba2007learning} applied an ILP learning algorithm to introduce the definitions of the two predicates and establish an ordering relationship. After that, \citep{malerba2008machine} investigated the problem of detecting the reading order relationship between components of a logical structure with domain specific knowledge.~\citep{10.1145/2644866.2644883} presented an unsupervised strategy for identifying the correct reading order of a document page's components based on abstract argumentation. The method is based on an empirical assumption about how humans behave when reading documents. More recently, deep learning models have become the mainstream solution for many machine learning problems. ~\cite{10.1007/978-3-030-58595-2_6} proposed an end-to-end OCR text reorganizing model, where they use a Graph Neural Network with an attention map to encode the text blocks with visual layout features, with an attention-based sequence decoder to reorder the OCR text into a proper sequence. 

\section{Conclusion}

In this paper, we present ReadingBank, a benchmark dataset for reading order detection that contains 500,000 document images. In addition, we also propose LayoutReader, a novel reading order detection approach built upon the pre-trained LayoutLM model. Experiments show that the LayoutReader has significantly outperformed the left-to-right and top-to-bottom heuristics as well as several strong baselines. Furthermore, the LayoutReader can be easily adapted to any OCR engines so that the reading order can be improved for downstream tasks. The ReadingBank dataset and LayoutReader model will be publicly available to support more research on reading order detection.

For future research, we will investigate how to generate a larger synthesized dataset from the ReadingBank, where noisy information and rotation can be applied to the clean images to make the model more robust. Moreover, we will label the reading order information on a real-world dataset from scanned documents. Considering the LayoutReader model as a pre-trained reading order detection model, we will also explore whether a few human labeled samples would be sufficient for the reading order detection in a specific domain.

\appendix

\section{Ethical Consideration}
The ethical impact of our research has always been an important consideration. While pursuing better performance and high quality datasets, we respect the intellectual property of the data resources. We sincerely hope our research will benefit the academia and foster more related study and, meanwhile, all ethical standards are strictly followed.

When building the new dataset, ReadingBank, we carefully crawl the public available data from the internet. We strictly follow the robots exclusion standard of each website to make sure we are permitted to collect the data. We also exclude the web pages with privacy issues and only keep those pages we have the permission to edit and redistribute according to the license rules. To guarantee there is no potential ethical violation, we will publicize a proportion of our dataset (about 100 pages) and this subset will be manually checked and redacted 
while the access of the whole version requires our further permission. All the data in our dataset will be protected by Apache 2.0 license.

We design the reading order detection as a fundamental task for the document image understanding. Numerous following tasks can be built on the basis of it. We do not set preference or limitation about the areas when we crawl the data so we believe the result of LayoutReader can be well generalized to other visually-rich document images due to the vast scope our dataset covers.

\bibliography{ref, anthology}
\bibliographystyle{acl_natbib}

\end{document}